\documentclass[10pt,twocolumn,letterpaper]{article}

\usepackage{wacv}
\usepackage{times}
\usepackage{epsfig}
\usepackage{graphicx}
\usepackage{amsmath}
\usepackage{amssymb}
\usepackage{booktabs}
\usepackage{enumitem}
\usepackage{xcolor}
\usepackage{subcaption}
\usepackage{pifont}
\usepackage[accsupp]{axessibility}

\definecolor{myOrange}{HTML}{ffae42}

\wacvalgorithmstrack   
\wacvfinalcopy 

\ifwacvfinal
\fi

\ifwacvfinal
\usepackage[breaklinks=true,bookmarks=false]{hyperref}
\else
\usepackage[pagebackref=true,breaklinks=true,colorlinks,bookmarks=false]{hyperref}
\fi

\begin{document}

\title{Holistic Interaction Transformer Network for Action Detection}


\author{Gueter Josmy Faure$^1$ \quad
Min-Hung Chen$^2$ \quad
Shang-Hong Lai$^{1,2}$ \\
$^1$National Tsing Hua University, Taiwan \quad
$^2$Microsoft AI R\&D Center, Taiwan \\
{\tt\small  josmyfaure@gapp.nthu.edu.tw \quad vitec6@gmail.com \quad lai@cs.nthu.edu.tw}
}

\maketitle
\thispagestyle{empty}

\begin{abstract}
   Actions are about how we interact with the environment,
including other people, objects, and ourselves. In this paper, we propose a novel multi-modal \textit{\textbf{H}olistic \textbf{I}nteraction \textbf{T}ransformer Network (\textbf{HIT})} that leverages the largely ignored, but critical hand and pose information  essential to most human actions.
The proposed \textbf{HIT} network is a comprehensive bi-modal framework that comprises an RGB stream and a pose stream. Each of them separately models person, object, and hand interactions. Within each sub-network, an Intra-Modality Aggregation module (IMA) is introduced that selectively merges individual interaction units. The resulting features from each modality are then glued using an Attentive Fusion Mechanism (AFM). Finally, we extract cues from the temporal context to better classify the occurring actions using cached memory. Our method significantly outperforms previous approaches on the J-HMDB, UCF101-24, and MultiSports datasets. We also achieve competitive results on AVA. The code will be available at \url{https://github.com/joslefaure/HIT}.
\end{abstract}

\section{Introduction}

\label{sec:intro}
\begin{figure}
  \centering
    \includegraphics[width=0.8\linewidth]{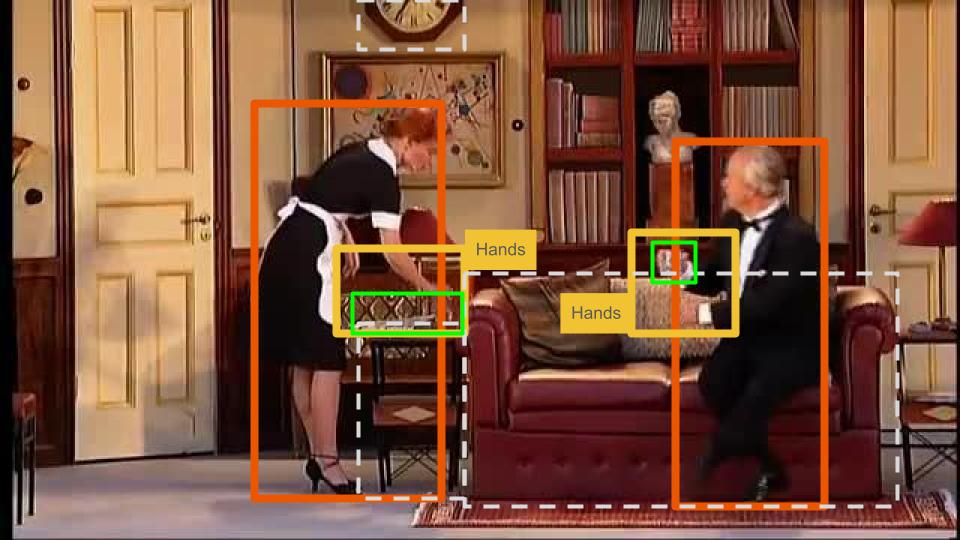}
    \caption{\textbf{Intuition}. This figure exemplifies how essential hand features are for detecting actions. Both persons in the frame are interacting with objects. Still, the instance detector fails to detect those very objects the persons are interacting with \textcolor{green}{(green boxes)} and, instead, picks the unimportant ones \textcolor{gray}{(dashed grey boxes)}. However, capturing the hands and everything in between \textcolor{myOrange}{(yellow boxes)} gives the model a better idea of the actions being performed by the actors \textcolor{red}{(red boxes)}; \textit{``lift/pick up"} (left) and \textit{``carry/hold"} (right).}
    \label{fig:intuition}
\end{figure}

Spatio-temporal action detection is the task of recognizing actions in space and in time. In this regard, it is fundamentally different and more challenging than plain action detection, whose goal is to label an entire video with a single class. A sound spatio-temporal action detection framework aims to deeply learn the information in each video frame to correctly label each person in the frame. It should also keep a link between neighboring frames to better understand activities with continuous properties such as  \textit{``open" - ``close"} \cite{carreira2017quo, feichtenhofer2019slowfast, ji20123d, qiu2017learning, tran2015learning}. In recent years, more robust frameworks have been introduced that explicitly consider the relationship between the spatial entities \cite{pan2021actor, wu2019long} since if two persons are in the same frame, they are likely to be interacting with each other. However, using only person features is insufficient for capturing object-related action (e.g., \textit{volleyball spiking}). Others try to understand the relationship not only between persons on the frame but also their surrounding objects \cite{materzynska2020something, tang2020asynchronous}. These methods have two main shortcomings. First, they only rely on objects with high detection confidence which might result in ignoring important objects that may be too small to be detected or unknown to the off-the-shelf detector. For example, in Figure \ref{fig:intuition}, none of the objects the actors are interacting with are detected. Secondly, these models struggle to detect actions related to objects not present in the frame. For instance, consider the action \textit{``point to (an object)"}. It is possible that the object the actor is pointing at is not in the current frame. \par

Figure \ref{fig:intuition} illustrates one of our motivations for undertaking this research. Most humans' actions are contingent on what they do with their hands and their poses when executing specific actions. The person on the left is \textit{``picking up/lifting (something)"} which is not noticeable even by humans. Still, our model is able to capture this action since we consider the person's hand features and the pose of the subject (the bending position is typical of someone picking up something). A similar issue occurs with the person on the right who is \textit{``sitting and holding (an object)"}. The man is holding a cup, but the object detector does not find the object, probably because it is very small or highly transparent. Using hand features, our model implicitly focuses on these challenging objects.

Our proposed \textbf{H}olistic \textbf{I}nteraction \textbf{T}ransformer (HIT) network uses fine-grained context, including person pose, hands, and objects, to construct a bi-modal interaction structure. Each modality comprises three main components: person interaction, object interaction, and hand interaction. Each of these components learns valuable local action patterns. We then use an Attentive Fusion Mechanism to combine the different modalities before learning temporal information from neighboring frames that help us better detect the actions occurring in the current frame. 
We perform experiments on the J-HMDB \cite{Jhuang:ICCV:2013}, UCF101-24 \cite{soomro2012ucf101}, Multisports \cite{li2021multisports} and AVA \cite{gu2018ava} datasets, and our method achieves state-of-the-art performance on the first three while being competitive with the SOTA methods on AVA. \par

\noindent The main contributions in this paper can be summarized as follows:
\begin{itemize}
  \item We propose a novel framework that combines RGB, pose and hand features for action detection.
  \item We introduce a bi-modal Holistic Interaction Transformer (HIT) network that combines different kinds of interactions in an intuitive and meaningful way.
  \item We propose an Attentive Fusion Module (AFM) that works as a selective filter to keep the most informative features from each modality and an Intra-Modality Aggregator (IMA) for learning useful action representations within the modalities.
  \item Our method achieves state-of-the-art performance on three of the most challenging spatio-temporal action detection datasets. 
\end{itemize}

\section{Related Work}
\label{sec:related}

\subsection{Video Classification}
Video classification consists in recognizing the activity happening in a video clip. Usually, the clip spans a few seconds and has a single label. Most recent approaches to this task use 3D CNNs~\cite{carreira2017quo, feichtenhofer2019slowfast, feichtenhofer2016convolutional, tran2015learning} since they can process the whole video clip as input, as opposed to considering it as a sequence of frames ~\cite{qiu2017learning, sun2015human}. Due to the scarcity of labeled video datasets, many researchers rely on models pre-trained on ImageNet ~\cite{carreira2017quo, wang2018non, xie2017rethinking} and use them as backbones to extract video features. Two-stream networks ~\cite{feichtenhofer2019slowfast, feichtenhofer2016convolutional} are another widely used approach to video classification thanks to their ability to only process a fraction of the input frames, striking a good balance between accuracy and complexity.

\subsection{Spatio-Temporal Action Detection}
In recent years, more attention has been given to spatio-temporal action detection \cite{feichtenhofer2019slowfast, girdhar2019video, li2019collaborative, pan2021actor, tang2020asynchronous}. As the name (spatio-temporal) suggests, instead of classifying the whole video into one class, we need to detect the actions in space, i.e., the actions of everyone in the current frame, and in time since each frame might contain different sets of actions. Most recent works on spatio-temporal action detection use a 3D CNN backbone ~\cite{ni2021identity, wu2019long} to extract video features and then crop the person features from the video features either using ROI pooling ~\cite{girshick2015fast} or ROI align ~\cite{he2017mask}. Such methods discard all the other potentially useful information contained in the video.


\subsection{Interaction Modeling}
What if the spatio-temporal action detection task really is an interaction modeling task? In fact, most of our everyday actions are interactions with our environment (e.g., other persons, objects, ourselves) and interactions between our actions (for instance, it is very likely that``open the door” is followed by ``close the door”). The interaction modeling idea spurs a wave of research about how to effectively model interaction for video understanding ~\cite{pan2021actor, tang2020asynchronous, wu2019long}. 

Most researches in this area use the attention mechanism. ~\cite{ma2018attend, zhou2018temporal} propose Temporal Relation Network (TRN), which learns temporal dependencies between frames or, in other words, the interaction between entities from adjacent frames. Other methods further model not just temporal but spatial interactions between different entities from the same frame ~\cite{materzynska2020something, tang2020asynchronous, wu2019long, yang2021beyond, zhou2021graph}. Nevertheless, the choice of entities for which to model the interactions differs by model. Rather than using only human features, ~\cite{pan2021actor, wu2020context} chose to use the background information to model interactions between the person in the frame and the context. They still crop the persons' features but do not discard the remaining background features. Such an approach provides rich information about the person's surroundings. However, while the context says a lot, it might induce noise. 

Attempting to be more selective about the features to use, ~\cite{materzynska2020something, tang2020asynchronous} first pass the video frames through an object detector, crop both the object and person features, and then model their interactions. This extra layer of interaction provides better representations than standalone human interaction modeling models and helps with classes related to objects such as \textit{``work on a computer"}. However, they still fall short when the objects are too small to be detected or not in the current frame.

\subsection{Multi-modal Action Detection}
Most recent action detection frameworks use only RGB features. The few exceptions such as \cite{gu2018ava, song2019tacnet, su2019improving, sun2018actor} and \cite{pramono2019hierarchical} use optical flow to capture motion. \cite{sun2018actor} employs an inception-like model and concatenates RGB and flow features at the $Mixed 4b$ layer (early fusion) whereas \cite{gu2018ava} and \cite{su2019improving} use an I3D backbone to separately extract RGB and flow features, then concatenate the two modalities just before the action classifier. While skeleton-based action recognition has been around for a while now \cite{chen2021channel, gupta2021quo, liu2020disentangling}, as far as we know, no previous works have tackled skeleton-based action detection. 

In this paper, we propose a bi-modal approach to action detection that employs visual and skeleton-based features. Each modality computes a series of interactions, including person, object, and hands, before being fused. A temporal interaction module is then applied to the fused features to learn global information regarding neighboring frames.

\section{Proposed Method}
\label{sec:method}
\begin{figure}
  \centering
    \includegraphics[width=0.9\linewidth]{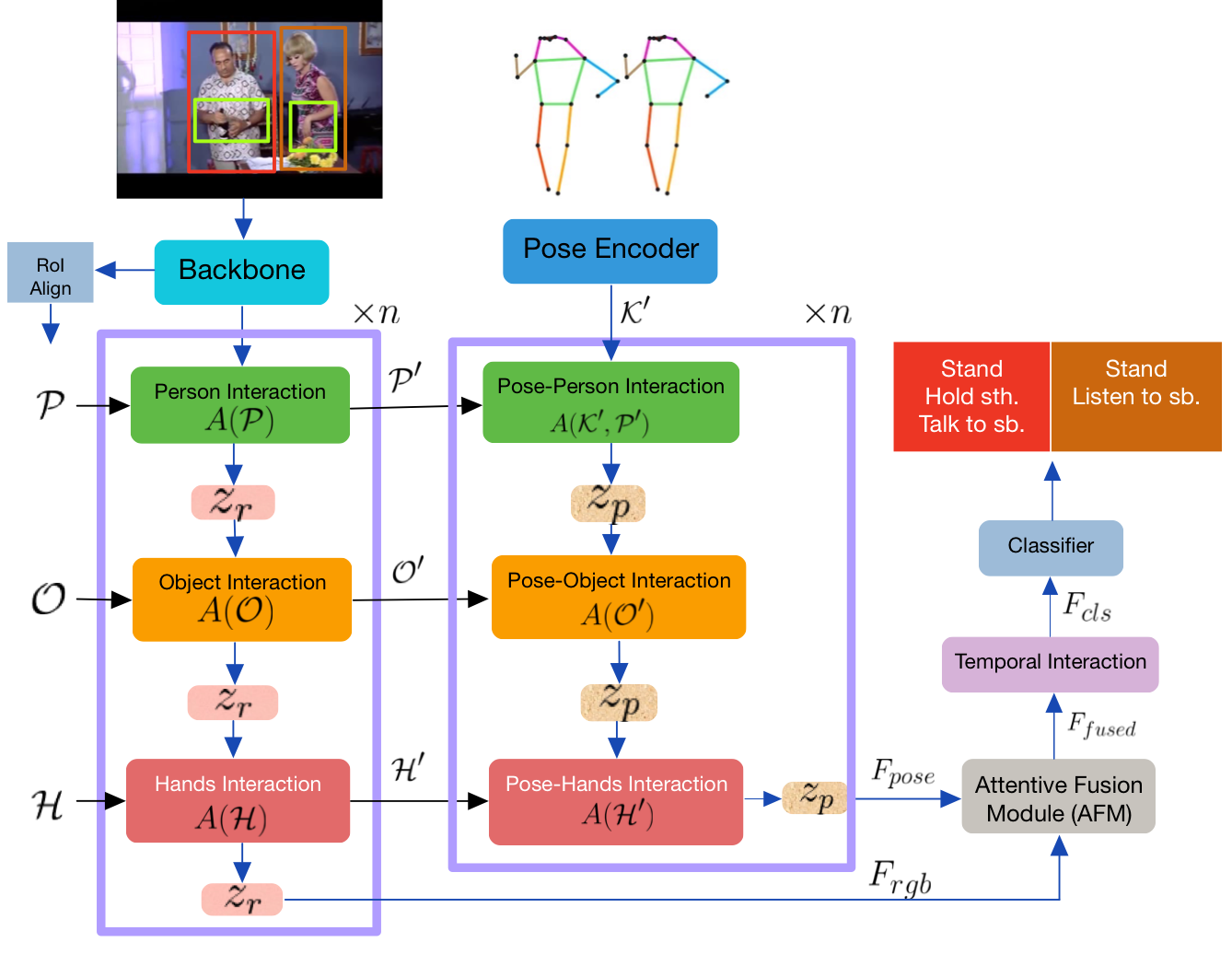}
    \caption{\textbf{Overview of our HIT Network.} On top of our RGB stream is a 3D CNN backbone which we use to extract video features. Our pose encoder is a spatial transformer model.
    We parallelly compute rich local information from both sub-networks using person, hands, and object features. We then combine the learned features using an attentive fusion module before modeling their interaction with the global context.}
    \label{fig:method}
\end{figure}

\begin{figure}
  \centering
    \includegraphics[width=0.5\linewidth]{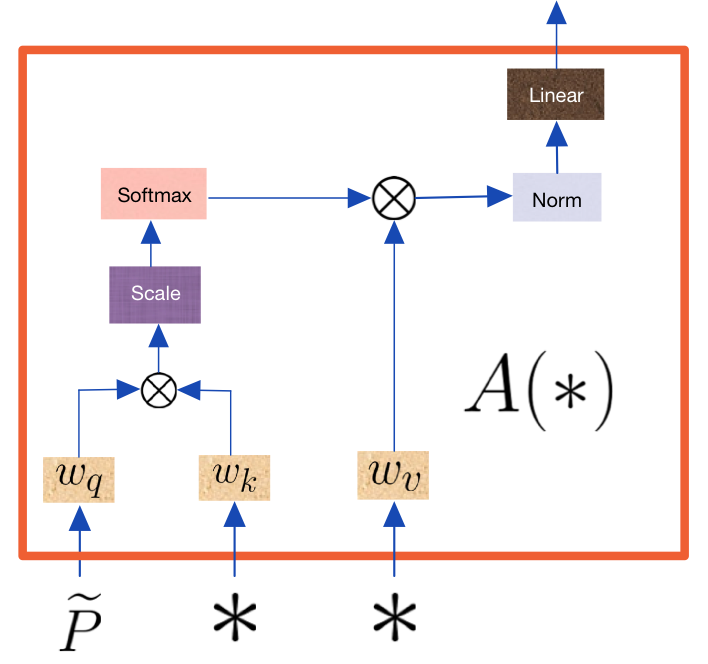}
    \caption{\textbf{Illustration of the Interaction module.} $*$ refers to the module-specific inputs while $\widetilde{P}$ refers to the person features in $A(\mathcal{P})$ or the output of the module that comes before $A(*)$.}
    \label{fig:interaction}
\end{figure}

\begin{figure}
  \centering
    \includegraphics[width=0.9\linewidth]{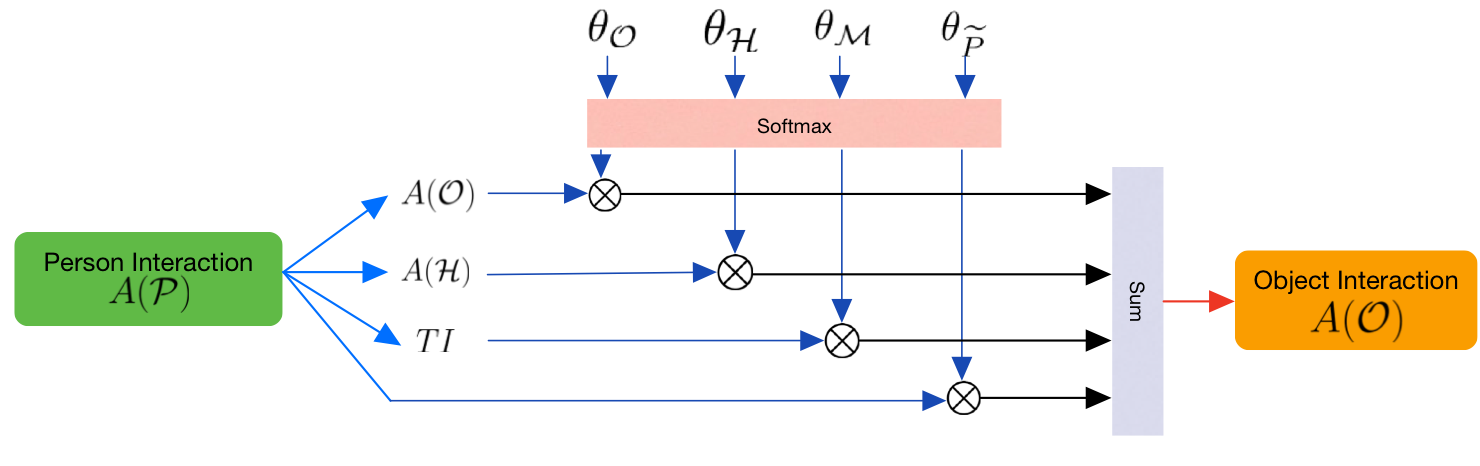}
    \caption{\textbf{Illustration of the Intra-Modality Aggregator.} Features from one unit to the next are first augmented with contextual cues then filtered.}
    \label{fig:ima}
\end{figure}

In this section, we provide a detailed walk-through of our approach. Our Holistic Interaction Transformer (HIT) network is concurrently composed of an RGB and a pose sub-network. Each aims to learn persons' interactions with their surroundings (space) by focusing on the key entities that drive most of our actions (e.g., objects, pose, hands). After fusing the two sub-networks' outputs, we further model how actions evolve in time by looking at cached features from past and future frames. Such a comprehensive activity understanding scheme helps us achieve superior action detection performance. 

This section is organized as follows:  we first describe the entity selection process in section \ref{subsec:instance}. In section \ref{subsec:rgb}, we elaborate on the RGB modality before introducing its pose counterpart in section \ref{subsec:pose}.  Further, in section \ref{subsec:fusion}, we explain our Attentive Fusion Module (AFM) and then the Temporal Interaction Unit (Section \ref{subsec:tiu}). 

Given an input video $V_{in} \in \mathbb{R}^{C \times T \times H \times W}$ we extract video features $V_b \in \mathbb{R}^{C \times T \times H \times W}$ by applying a 3D video backbone. Afterward, using ROIAlign, we crop person features $\mathcal{P}$, object features $\mathcal{O}$, and hands features $\mathcal{H}$ from the video. We also keep a cache of memory features $\mathcal{M} = [t-S, ..., t-1, t, t+1, ..., t+S]$, where $2S+1$ is the temporal window. Parallelly, we use a pose model to extract person keypoints $\mathcal{K}$ from each keyframe of the dataset. Further, the RGB and pose sub-networks compute the RGB feature ${F}_{rgb}$ and pose feature ${F}_{pose}$, respectively. These features are then fused and subsequently used as anchors for learning global context information to obtain $F_{cls}$. Finally, our network outputs $\hat{y}=g(F_{cls})$, where $g$ is the classification head. The overall framework is shown in Figure \ref{fig:method}.

\subsection{Entity Selection}
\label{subsec:instance}
HIT consists of two mirroring modalities with distinct modules designed to learn different types of interactions. Human actions are largely based on their pose, hand movements (and pose), and interaction with their surroundings. Based on these observations, we select human poses and hands bounding boxes as entities for our model, along with object and person bounding boxes. We use Detectron ~\cite{girshick2011detectron} for human pose detection and create a bounding box encircling the location of the person's hands. Following the state-of-the-art methods, \cite{tang2020asynchronous}, \cite{seong2019video}, \cite{pan2021actor}, we use Faster-RCNN \cite{ren2015faster} to compute object bounding box proposals. The video feature extractor is a 3D CNN backbone network\cite{feichtenhofer2019slowfast}, and the pose encoder is a lightweight spatial transformer inspired by \cite{zheng20213d}. We apply ROIAlign \cite{he2017mask} to trim the video features and extract person, hands, and objects features.


\subsection{The RGB Branch}
\label{subsec:rgb}
The RGB branch comprises three main components, as shown in Figure \ref{fig:method}. Each performs a series of operations to learn specific information concerning the target person. The person interaction module learns the interaction between persons in the current frame (or self-interaction when the frame contains only one subject). The object and hands interaction modules model person-object and person-hands interaction, respectively. At the heart of each interaction unit is a cross-attention computation where the query is the target person (or the output of the previous unit), and the key and value are derived from the objects, or the hands features, depending on which module we are at (see figure \ref{fig:interaction}). It is like asking ``how can these particular features help detect what the target person is doing?". The following equations summarize the RGB branch's flow.

\begin{equation}
\label{eq:rgb}
\begin{aligned}
F_{rgb}=(A(\mathcal{P}) \rightarrow z_r \rightarrow A(\mathcal{O}) \rightarrow z_r \rightarrow A(\mathcal{H}) \rightarrow z_r) \\ 
A(*) = softmax(\frac{w_q(\widetilde{P}) \times w_k(*)}{\sqrt{d_r}}) \times w_v(*) \\ 
z_r = \sum_{b} {A(b) \times softmax(\theta_{b})}, b \in{(\widetilde{P}, \mathcal{O}, \mathcal{H}, \mathcal{M})}
\end{aligned}    
\end{equation}
$d_r$ represents the channel dimension of the RGB features, $w_q$, $w_k$ and $w_v$ project their inputs into query, key and value, respectively. $A(*)$ is the cross-attention mechanism. It only takes person features as input when computing person interaction $A(\mathcal{P})$. However, for hand interaction (objects interaction), it takes two sets of input: the output of $z_r$, which serves as query (denoted as $\widetilde{P}$), and the hands features (object features) from which we obtain the key and values.  

The intra-modality aggregation component, $z_r$ is the weighted sum of all interaction modules, including the temporal interaction module $TI$ (see Figure \ref{fig:ima}). $z_r$ is essential for two main reasons. First, it allows the network to aggregate as much information as possible, efficiently. Secondly, the learnable parameter $\theta$ helps filter the different sets of features, hand-picking the best each of them has to offer while discarding noisy and unimportant information. A more detailed discussion on $z_r$ is provided in the supplementary material.

\subsection{The Pose Branch}
\label{subsec:pose}
The pose model is similar to its RGB counterpart and reuses most of its outputs. We first extract the pose features $\mathcal{K\sp{\prime}}$ by using a light transformer encoder $f$ inspired by \cite{zheng20213d}.
\begin{equation}
\mathcal{K\sp{\prime}} = f(\mathcal{K})
\end{equation}
Then we compute $F_{pose}$ by mirroring the different constituents of the RGB modality and reusing their corresponding outputs. Here, $\mathcal{P\sp{\prime}}$, $\mathcal{O\sp{\prime}}$, and $\mathcal{H\sp{\prime}}$ are the corresponding outputs of $A(\mathcal{P})$, $A(\mathcal{O})$, and, $A(\mathcal{H})$. 
\begin{equation}
\label{eq:pose}
\begin{aligned}
F_{pose}=(A(\mathcal{K\sp{\prime}}, \mathcal{P}\sp{\prime}) \rightarrow z_p \rightarrow A(\mathcal{O}\sp{\prime}) \rightarrow z_p \rightarrow A(\mathcal{H}\sp{\prime}) \rightarrow z_p) \\
A(\mathcal{K\sp{\prime}}, \mathcal{P}\sp{\prime}) = softmax(\frac{w_q(\mathcal{K\sp{\prime}}) \times w_k(\mathcal{P}\sp{\prime})}{\sqrt{d_p}}) \times w_v(\mathcal{P}\sp{\prime})
\end{aligned}    
\end{equation}

$A(\mathcal{K\sp{\prime}}, \mathcal{P}\sp{\prime})$ computes the cross-attention between the pose features $\mathcal{K\sp{\prime}}$ and the enhanced person interaction features $\mathcal{P}\sp{\prime}$. Such a cross-modal blend enforces the pose features by focusing on the key corresponding attributes of the RGB features. The other components, $A(\mathcal{O}\sp{\prime})$ and $A(\mathcal{H}\sp{\prime})$ take a linear projection of $z_p$ as query while their key-value pairs stem from $A(\mathcal{O})$ and $A(\mathcal{H})$. $z_p$ is the intra-modality aggregation component for the pose model. Similar to $z_r$, it filters and aggregates information from each interaction module.

\subsection{The Attentive Fusion Module (AFM)}
\label{subsec:fusion}
At some point in the network, the RGB and pose streams need to be combined into one set of features before being fed to the action classifier. For this purpose, we propose an Attentive Fusion Module that applies channel-wise concatenation of the two feature sets followed by self-attention for feature refinement. We then reduce the magnitude of the output feature by using the projection matrix $\Theta_{fused}$. Table \ref{tab:afm} in our ablation study validates the superiority of our fusion mechanism compared to other fusion types used in the literature.
\begin{equation}
\label{eq:fusion}
\begin{aligned}
F_{fused} = \Theta_{fused}(SelfAttention(F_{rgb}, F_{pose}))
\end{aligned}
\end{equation}

\subsection{Temporal Interaction Unit}
\label{subsec:tiu}
Following the fusion module is a temporal interaction block $(TI)$. Human actions happen in a continuum; therefore, long-term context is essential to understanding actions. Along with $F_{fused}$, this modules receives compressed memory data $\mathcal{M}$ with length $2S+1$. Inspired by \cite{tang2020asynchronous}, the memory cache contains the person features extracted by the video backbone. $F_{fused}$ inquires $\mathcal{M}$ as to which of the neighboring frames contains informative features, then absorbs them. $TI$ is another cross-attention module where $F_{fused}$ is the query and two different projections of the memory $\mathcal{M}$ form the key-value pair. 
\begin{equation}
\label{eq:cls}
\begin{aligned}
F_{cls} = TI(F_{fused}, \mathcal{M}) \\
\end{aligned}    
\end{equation}

Finally, the classification head $g$ is composed of two feed-forward layers with relu activation, and the output layer.
\begin{equation}
\label{eq:yhat}
\begin{aligned}
\hat{y}=g(F_{cls})
\end{aligned}    
\end{equation}


\section{Experiments}
We perform experiments on four challenging action detection datasets: J-HMDB \cite{Jhuang:ICCV:2013}, UCF101-24 \cite{soomro2012ucf101}, MultiSports \cite{li2021multisports} and AVA \cite{gu2018ava}. The implementation details described below relate to the J-HMDB and UCF101-24 datasets. We refer the reader to the supplementary materials for details on how we train MultiSports and AVA.\\
\subsection{Datasets}
The \textbf{J-HMDB dataset} \cite{Jhuang:ICCV:2013} has 21 action classes and up to 55 clips per class. The dataset totaled 31,838 annotated frames with a resolution of 320x240. Each video clip is trimmed to contain a single action. To be on the same page with other methods, we report frame and video mAP results on split-1 of the dataset. The IoU threshold for frame mAP is 0.5, the same as other methods in our comparison table. 

\textbf{UCF101-24} is a subset of the UCF101 \cite{soomro2012ucf101} dataset suitable for Spatio-temporal action detection. It contains 24 action classes (mainly related to sports activities) spanning 3207 untrimmed videos with human bounding boxes annotated frame-by-frame. We employ the same testing protocol as that of J-HMDB.

\textbf{MultiSports} \cite{li2021multisports} contains 66 fine-grained action categories from four different sports spanning more than 3200 video clips with 37701 action instances and 902k bounding boxes. Actions are annotated at 25 FPS, and each video clip lasts around 22 seconds.

\textbf{AVA} \cite{gu2018ava} version 2.2 consists of 430 15-minutes videos sampled from YouTube. For each video in the dataset, 900 frames are annotated with human bounding boxes and labels. The dataset contains 80 class labels divided into pose action (14), person-person interaction (49), and person-object interaction (17) classes. Following the standard practice, we report the frame mAP for 60 of the 80 classes with a spatial IoU threshold of 0.5.

\subsection{Implementation Details}
\noindent{\textbf{Person and Object Detector:}} We extract keyframes from each video in the dataset and use detected person bounding boxes from \cite{kopuklu2019you} for inference. As object detector, we employ Faster-RCNN \cite{ren2015faster} with ResNet-50-FPN \cite{lin2017feature, xie2017aggregated} backbone. The model is pretrained on ImageNet \cite{deng2009imagenet}, and fine-tuned on MSCOCO \cite{lin2014microsoft}. 

\noindent{\textbf{Keypoints Detection and Processing:}} For keypoints detection, we adopt a pose model from Detectron ~\cite{girshick2011detectron}. The authors use a Resnet-50-FPN backbone pretrained on ImageNet for object detection and fine-tuned on MSCOCO keypoints using precomputed RPN \cite{ren2015faster} proposals. Each keyframe from the target dataset is passed through the model, which outputs 17 keypoints for each detected person, corresponding to the COCO format. We further post-process the detected pose coordinates, so they match the groundtruth person bounding boxes (during training) and the bounding boxes from \cite{kopuklu2019you} (during testing). For person hands location, we are only interested in the keypoints referring to the person’s wrists; therefore, we make a bounding box out of these two keypoints to highlight the person’s hands and everything in between. 

\noindent{\textbf{Backbone:}} We employ SlowFast networks \cite{feichtenhofer2019slowfast} as our video backbone. Our experiments and ablation study use SlowFast with a ResNet-50 instantiation pretrained on Kinetics-700 \cite{carreira2017quo}. For AVA and MultiSports, we use the more powerful SlowFast-Resnet-101 pretrained on K700.

\noindent{\textbf{Training and Evaluation:}} The input videos are sampled 32 frames per clip, with $\alpha = 4$ and $\tau = 1$, meaning the SlowFast backbone has a temporal stride of 4 for the slow path while the fast path takes the entire 32 frames as input. During training, random jitter augmentation is applied to the ground-truth human bounding boxes. For object boxes, we use the ones with detection score $\geq 0.25$ and whose $IoU$ with any person bounding box in the same frame is positive. This is to ensure that only the objects with relatively high confidence scores and those with which humans directly interact are included in our sample. The network is trained on the J-HMDB dataset for 7K iterations, with the first 700 iterations serving as linear warm-up. No weight decay was used. We use SGD as optimizer and a batch size of 8 to train the model on one 11GB GPU. We train on the UCF101-24 dataset for 50k iterations, adopting linear warm-up during the first 1k iterations. The starting learning rate of 0.0002 is reduced by a factor of 10 at iterations 25k and 35k. During inference, we predict action labels for human bounding boxes provided by \cite{kopuklu2019you} for both datasets.

\begin{table}[ht]
 \centering
\begin{tabular}{lc|c|cc}
\toprule
    Model & input & f@0.5 & v@0.2 & v@0.5 \\
    \toprule
    ACT \cite{kalogeiton2017action} & V + F & 65.7 & 74.2 & 73.7 \\
    Li et. al~\cite{li2020finding} & V & --- & 76.1 & 74.3 \\
    TacNet~\cite{song2019tacnet} & V + F & 65.5 & 74.1 & 73.4 \\
    MOC~\cite{li2020actions} & V + F & 70.8 & 77.3 & 77.2 \\
    AVA~\cite{gu2018ava}  & V + F & 73.3 & --- & 78.6 \\
    PCSC~\cite{su2019improving} & V + F & 74.8 & 82.6 & 82.2 \\
    HISAN~\cite{pramono2019hierarchical} & V + F & 76.7 & 85.9 & 84.0 \\
    ACRN~\cite{sun2018actor} & V + F & 77.9 & --- & 80.1 \\
    Context rcnn~\cite{wu2020context} & V & 79.2 & --- & --- \\
    TubeR~\cite{zhao2022tuber} & V + F & --- &  87.4 & 82.3 \\
    \midrule
    \textbf{Ours} & V & \textbf{83.8} & \textbf{89.7} & \textbf{88.1} \\
    \bottomrule
  \end{tabular}\hfill%
  \caption{\textbf{Frame and video-level comparison with the state-of-the-art methods on J-HMDB.} We use a SlowFast-Resnet50 as video backbone and report our results in mAP. Our model outperforms state-of-the-art methods on both frame mAP and video mAP metrics.}
    \label{tab:jhmdb}
\end{table}

\begin{table}[ht]
 \centering
\begin{tabular}{lc|r|cc}
\toprule
    Model & input & f@0.5 & v@0.2 & v@0.5 \\
    \toprule
    ACT \cite{kalogeiton2017action} & V + F & 67.1 & 77.2 & 51.4 \\
    ACDNet \cite{liu2021acdnet} & V + F & 70.9 & --- & --- \\
    TacNet \cite{song2019tacnet} & V + F & 72.1 & 77.5 & 52.9 \\
    HISAN \cite{pramono2019hierarchical} & V + F & 73.7 & 80.4 & 49.5 \\
    MOC \cite{li2020actions} & V + F & 78.0 & 82.8 & 53.8 \\
    AIA \cite{tang2020asynchronous} & V & 78.8 & --- & --- \\
    PCSC \cite{su2019improving} & V + F & 79.2 & 84.3 & 61.0 \\
    TubeR \cite{zhao2022tuber} & V + F & 83.2 & 83.3 & 58.4 \\
    ACAR \cite{pan2021actor} & V & 84.3 & --- & ---\\
    \midrule
    \textbf{Ours} & V & \textbf{84.8}  & \textbf{88.8} & \textbf{74.3} \\
    \bottomrule
  \end{tabular}\hfill%
  \caption{\textbf{Comparison with the State-of-the-art methods on UCF101-24}. Like other methods in our comparison table, we evaluate frame mAP on split 1 with an $IoU$ threshold of 0.5 and video mAP with thresholds of 0.2 and 0.5.}
    \label{tab:ucf}
\end{table}

\begin{table}[ht]
 \centering
\begin{tabular}{l|c|cc}
\toprule
    Model & f@0.5 & v@0.2 & v@0.5 \\
    \toprule
    ROAD \cite{singh2017online} & 3.9 & 0.0  & 0.0 \\
    YOWO\cite{kopuklu2019you} & 9.2 & 10.7 & 0.8 \\
    MOC \cite{li2020actions} & 25.2 & 12.8 & 0.6 \\
    MultiSports \cite{li2021multisports} & 27.7 & 24.1 & \textbf{9.6} \\
    \midrule
    Ours & \textbf{33.3} & \textbf{27.8} & 8.8 \\
    \bottomrule
  \end{tabular}\hfill%
  \caption{\textbf{Comparison with the State-of-the-art on MultiSports.} Our model significantly outperforms the other methods on two metrics.}
  \label{tab:multisports_res}
\end{table}

\begin{table}[ht]
 \centering
\begin{tabular}{l|c|c}
\toprule
    Model & Pretrain & frame mAP \\
    \toprule
    SlowFast, R-101+NL \cite{feichtenhofer2019slowfast} & K600 & 29.0 \\
    X3D-L\cite{feichtenhofer2020x3d} & K600 & 29.4 \\
    AIA \cite{tang2020asynchronous} & K700 & 32.3 \\
    Object Transformer\cite{wu2021towards} & K600 & 31.0 \\
    Beyond Short Clips\cite{yang2021beyond} & K700 & 31.6 \\
    ACAR\cite{pan2021actor} & K700 & 33.3\\
    MeMViT \cite{wu2022memvit} & K700 & 33.5 \\
    *TubeR \cite{zhao2022tuber} & IG + 400 & \textbf{33.6} \\
    \midrule
    Ours & K700 &  32.6 \\
    \bottomrule
  \end{tabular}\hfill%
  \caption{\textbf{Comparison with the State-of-the-art on AVA v2.2.} Our model has comparable results compared to the SOTA methods.}
  \label{tab:avares}
\end{table}

\subsection{Comparison with State-of-the-Art Methods}
In Tables \ref{tab:jhmdb} and \ref{tab:ucf}, we compare our results with other methods on the challenging J-HMDB and UCF101-24 datasets, respectively. Our method registers significant gains compared to the state-of-the-art methods both in terms of frame and video mAP. Such a performance demonstrates our bi-modal framework's ability to capture more diverse clues about human actions by taking a closer look at the human's pose and environment.

In Table \ref{tab:multisports_res}, we report our results on the MultiSports dataset. Our method outperforms other methods in terms of frame mAP with an IoU threshold of 0.5, and video mAP when the spatio-temporal tube threshold is 2. As Table \ref{tab:avares} shows, we achieve competitive results on the most challenging fine-grained action detection dataset (AVA). With ACAR \cite{pan2021actor} using pretrained features as memory and TubeR \cite{zhao2022tuber} using a backbone pretrained on the IG + K400 dataset, the only comparable method that outperforms ours is MeMViT \cite{wu2022memvit}. Overall, our results on four action detection datasets exhibit the generalization capabilities of our method.

\begin{table*}
 \centering
 \label{tab:ablation}
\begin{subtable}[t]{0.25\linewidth}\centering
\begin{tabular}[t]{@{}lc@{}}
    \toprule
    Bi-modal fusion & mAP \\
    \midrule
    Sum & 78.60 \\
    Concat & 78.77 \\
    WeightedSum & 80.21 \\
    Average & 81.35 \\
    AFM & \textbf{83.81} \\
    \bottomrule
  \end{tabular}
  \caption{\textbf{Bi-modal fusion methods}}
  \label{tab:afm}
\end{subtable}\hfill%
\begin{subtable}[t]{0.2\linewidth}\centering
\begin{tabular}[t]{@{}lc@{}}
    \toprule
    Depth & mAP \\
    \midrule
    1 layer & 79.21 \\
    2 layers & \textbf{83.81} \\
    3 layers & 81.54 \\
    \bottomrule
  \end{tabular}
  \caption{\textbf{Network Depth}}
  \label{tab:depth}
\end{subtable}\hfill%
\begin{subtable}[t]{0.3\linewidth}\centering
\begin{tabular}[t]{@{}lc@{}}
    \toprule
     & mAP \\
    \midrule
    After Temporal Interaction & 82.16 \\
    Before Temporal Interaction & \textbf{83.81} \\
    \bottomrule
  \end{tabular}
\caption{\textbf{Late versus early fusion}}
\label{tab:fusion}
\end{subtable}\hfill%
\begin{subtable}[t]{0.2\linewidth}\centering
\begin{tabular}[t]{@{}lc@{}}
    \toprule
     & mAP \\
    \midrule
    w/o IMA & 79.80 \\
    w/ IMA  & \textbf{83.81} \\
    \bottomrule
  \end{tabular}
\caption{\textbf{Importance of IMA}}
\label{tab:intra}
\end{subtable}\hfill%

\begin{subtable}[b]{0.3\linewidth}\centering
\begin{tabular}[t]{@{}lc@{}}
    \toprule
     & mAP \\
    \midrule
    Backbone &  58.85 \\
    Backbone + AIA\cite{tang2020asynchronous} & 77.25 \\
    Backbone + Pose Encoder & 80.44 \\
    Backbone + Ours & \textbf{83.81} \\
    \bottomrule
  \end{tabular}
\caption{\textbf{Interaction modeling methods}}
\label{tab:interaction}
\end{subtable}\hfill%
\begin{subtable}[b]{0.3\linewidth}\centering
\begin{tabular}[t]{llcr}
    \toprule
    Hands & RGB & Pose & mAP \\
    \midrule
     &  &  & 58.85 \\
    \midrule
     & \checkmark &  & 79.11 \\
     & \checkmark & \checkmark & 79.62 \\
     &  & \checkmark & 80.19 \\
    \checkmark  & \checkmark &  & 80.82 \\
    \checkmark &  & \checkmark & 80.90 \\
    \midrule
    \checkmark  & \checkmark & \checkmark & \textbf{83.81} \\
    \bottomrule
  \end{tabular}
  \caption{\textbf{The importance of each modality and the hand features}}
  \label{tab:hands}
\end{subtable}\hfill%
\begin{subtable}[b]{0.3\linewidth}\centering
\begin{tabular}[t]{llcr}
    \toprule
    $A(\mathcal{H})$ & $A(\mathcal{O})$ & $TI$ & mAP \\
    \midrule
    \checkmark & & & \underline{81.44} \\
    & \checkmark & & 78.86 \\
    & & \checkmark & 79.73 \\
    \checkmark & & \checkmark & 79.36 \\
    \checkmark & \checkmark & & 80.23 \\
    & \checkmark & \checkmark & 79.62 \\
    \midrule
    \checkmark  & \checkmark & \checkmark & \textbf{83.81} \\
    \bottomrule
  \end{tabular}
\caption{\textbf{Importance of individual interaction units}}
\label{tab:single_unit}
\end{subtable}\hfill%
\caption{\textbf{Ablation Study on J-HMDB} We use a SlowFast-Resnet50 as video backbone and report our results in mAP. \textit{backbone} refers to the video \textit{backbone} followed by the action classifier. For \textit{Backbone + Encoder} we directly use our AFM to fuse the pose and RGB features extracted from the pose encoder and video backbone, then apply the action classifier.}
\end{table*}

\subsection{Ablation Study}
\label{subsec:ablation}
We perform ablation experiments on the J-HMDB dataset to illustrate the effectiveness of our model and its constituents. All ablations are performed using the SlowFast-Resnet50 video backbone. We use frame mAP with an IoU threshold of 0.5 as evaluation metric.

\noindent{\textbf{Network Depth:}} Two layers of our network are enough to learn valuable features conducing to accurate action detection. As shown in Table \ref{tab:depth}, a two-layer setting improves the mAP by more than $4\%$ compared to having just one, while adding a third induces overfitting. This is due to our method blending a lot of information within one layer. Therefore, for the remaining experiments, we report results using two layers. By two layers, we mean that the RGB sub-network is repeated twice, and so is the pose sub-network.

\noindent{\textbf{Attentive Fusion Module (AFM):}}
We used an Attentive Fusion Mechanism (AFM) to combine features from the two modalities. Equipped with self-attention, it helps smoothen the fusion process between different modalities. We corroborate this choice by comparing it with $Sum$, $Concat$, $Weighted Sum$, and $Average$. 

In the $Sum$ fusion, refers to element-wise addition of the features. Such a method yields the worst result since we end up with significantly magnified result. The $Concat$ fusion stands for channel-wise concatenation of the RGB and pose features. It is slightly better than the $Sum$ fusion but still falls short of the desired outcome since it does not enhance the results. $Weighted Sum$ yields a marginally higher mAP than the two previous fusion methods. However, it does not challenge our AFM since our intra-modality aggregator (IMA: $z_r$, $z_p$) already selects the best features from each modality. A better fusion method is the $Average$ fusion, which takes the average of the RGB and pose streams. Such a fusion approach solves the shortcomings of $Sum$ but does not enhance the resulting feature. As shown in table \ref{tab:afm}, our AFM works better than the other approaches by virtue of its ability to enhance the combined features.

\noindent{\textbf{Late vs. Early Fusion:}} Late/early fusion refers to whether we fuse the two modalities before or after the Temporal Interaction module. Table \ref{tab:fusion} reports our results trying both structures. As we expected, temporal interaction works best when it's done on the full feature map, instead of features from each modality independently. It should also be more efficient since we only need one temporal interaction unit.

\noindent{\textbf{The Intra-Modality Aggragator (IMA):}}
In section \ref{sec:method}, we describe the use of the intra-modality component $z_r$ for the RGB modality and $z_p$ for the pose model. We notice that better feature selection is achieved when the network learns by itself how to do that. As shown in Table \ref{tab:intra}, without the intra-modality aggregation module, important information would be wasted, holding back the model's performance. Therefore, we present the features from each interaction unit and let the IMA component choose and aggregate information as it pleases.

\begin{figure}[ht]
\centering
        \begin{subfigure}[t]{.23\textwidth}
            \centering
            \captionsetup{justification=centering}
            \includegraphics[width=.9\linewidth]{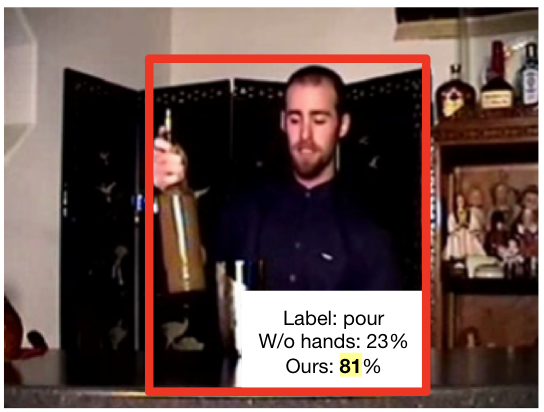}
            \caption{Hand features is essential for detecting the action class ``pour".}
            \label{fig:sfig1}
        \end{subfigure}%
        \begin{subfigure}[t]{.23\textwidth}
            \centering
            \includegraphics[width=.9\linewidth]{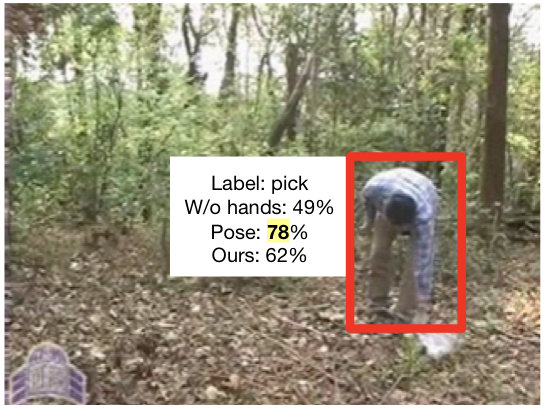}
            \caption{The ``pick up" class has a clear pose signature.}
            \label{fig:sfig2}
        \end{subfigure}
        \caption{\textbf{On the hand and pose features importance.} }
        \label{fig:qualitative1}
\end{figure}

\begin{figure}[ht]
\centering 
        \begin{subfigure}[t]{.25\textwidth}
            \centering
            \captionsetup{justification=centering}
            \includegraphics[width=.9\linewidth]{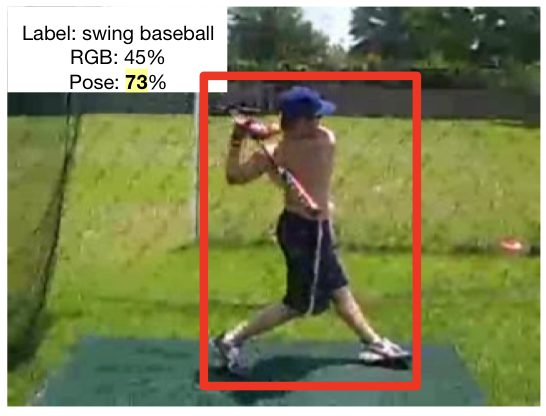}
            \caption{Another action with clear pose signature.}
            \label{fig:sfig3}
        \end{subfigure}%
        \begin{subfigure}[t]{.25\textwidth}
            \centering
            \captionsetup{justification=centering}
            \includegraphics[width=.9\linewidth]{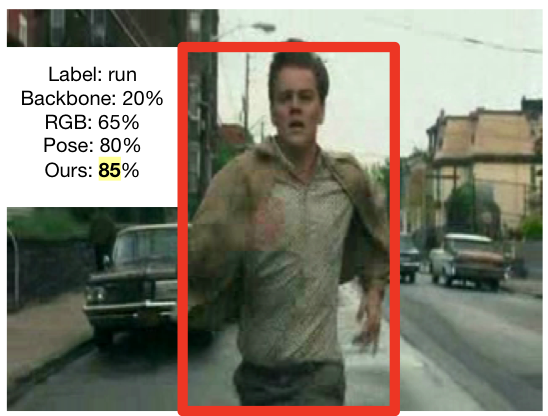}
            \caption{A neutral class. The accuracy increases as a we plug in more modules.}
            \label{fig:sfig4}
        \end{subfigure}%
        \caption{\textbf{The modalities' importance.} }
        \label{fig:qualitative2}
\end{figure}

\noindent{\textbf{Interaction Modeling methods:}}
To validate our interaction modeling scheme, we re-implement another interaction method found in the literature on top of the video backbone network. Table \ref{tab:interaction} contains results obtained with the bare backbone, with the backbone and our pose encoder, and the implementation of AIA \cite{tang2020asynchronous}. For the \textit{Backbone + Pose Encoder} framework, we directly fuse the outputs of the video backbone and the pose encoder. The table shows that our pose encoder is stronger than AIA, which aggregates person, object, and memory interaction. This proves that a person's pose contains rich information about what the person is doing. Such a result also confirms that pose information works well, whether used as a supplement or as a standalone network.

\noindent{\textbf{The importance of each modality and the hand features:}} In Table \ref{tab:hands}, we present a detailed ablation of the different building blocks of our model. Using only the RBG or pose modality, the action detection mAP jumps $20$ points compared to the backbone and keeps increasing from there. Hand features excluded, the pose-only model is stronger than the RGB-only model, which confirms our assumption that hand features are more valuable to the RGB sub-network since the pose sub-network implicitly contains hand information (hand keypoints). That being said, the pose-only model still benefits from hand features, as evidenced by the mAP increase from $80.19\%$ without hands to $80.90\%$ with them. The RGB-only model registers a higher gain when hands are added ($79.11\%$ versus $80.82\%$). These experiments underline the importance of hand interaction for action detection. With all these components pulling the strings, the model trained with both modalities with hand interaction registers the highest accuracy. Such an outcome indicates the harmony between all parts of our framework as well as their independent contributions.

\noindent{\textbf{Importance of Different Types of Interactions: }}
Since our framework is composed of three auxiliary types of interaction units, we wanted to quantify their different contributions. While it is feasible, we did not consider removing $A(\mathcal{P})$ in this ablation since our model is person-centric. As Table \ref{tab:single_unit} shows, hands interaction ($A(\mathcal{H})$) alone yields higher accuracy than either $A(\mathcal{O})$ or $TI$. It is also better than any other combination. We suspect this is a byproduct of our Intra-Modality Aggregator not having enough features to work with. Without other interaction types as enforcers, $A(\mathcal{O})$ returns the lowest accuracy. However, when paired with hand interaction, the model's accuracy jumps from 78.86\% to 80.23\%, outlining their complementarity. This ablation proves that the previously ignored hand features provide essential information for accurate action detection.

\subsection{Qualitative Results}

To further assess our framework's performance and understand what it ``sees", in Figure \ref{fig:qualitative1}, we present qualitative results on select frames from the J-HMDB dataset with action classes we consider as hand-related. Figure \ref{fig:sfig1} illustrates how using hand features can help for classes related to hands, such as ``pour". A model without hands would struggle to detect such an action due to the poor disparity between the background and the actor. Our model easily spots the action since it, among other things, focuses on the person's hand. In Figure \ref{fig:sfig2}, the pose-only model is even more powerful than the complete bi-modal framework due to the person's bending, which is a strong pose feature. Even though the action of ``picking up something" is hand-related, hand detection features for this frame might be noisy because of the blurriness of the frame. Such a result demonstrates the subtleties our pose modality is able to identify. 

Figure \ref{fig:sfig3} confirms that our pose-only model does an exceptional job classifying actions with typical pose signatures. The person uses his hand to ``swing a baseball"; however, the pose signature is still more evident than the RGB hand features. Figure \ref{fig:sfig4} further confirms the significance of each modality of our model. For a neutral class like ``run", the model's confidence keeps increasing as we add the modalities, reaching its peak with both RGB and pose combined. With such an outcome, we can argue that the different modalities of our network work in tandem to help us achieve superior video action detection performance.

\subsection{Limitations and failure cases}
Our framework depends on the off-the-shelf detector and pose estimator used and does not account for their failure. A large number of frames of the AVA dataset are crowded and have low quality. Therefore, the detector and pose estimator's accuracy might affect our method's. Analyzing our results on the J-HMDB dataset, we found two main causes of failure. The first relates to similar-looking classes, such as ``throw" and ``catch", which are visually identical. The second is partial occlusion. Please refer to the supplementary material for more thorough discussions on the limitations.

\section{Conclusion}
Learning the nature of interactions between person and other instances is vital for detecting actions. In this paper, we demonstrate that a careful choice of instances is crucial for a sound action detection framework. In our \textbf{H}olistic \textbf{I}nteraction \textbf{T}ransformer (HIT) Network, we integrate previously ignored entities such as person pose and hands and construct a bi-modal framework to model and aggregate interactions effectively. Modality-specific interaction features are combined using our proposed Attentive Fusion Mechanism. We also present detailed ablations validating our design choices. Results on four public action detection benchmarks demonstrate our framework's superiority over state-of-the-art methods.

\noindent{\textbf{Acknowledgements:}}
This research was supported in part by National Science and Technology Council, Taiwan, under the grant 111-2221-E-007-106-MY3. We thank Wei-Jhe Huang for constructive discussions.

\pagebreak
\begin{center}
\textbf{\large Supplemental Materials}
\end{center}
\setcounter{section}{0}


In the supplementary material, we aim to provide deeper technical details about our framework, and show more qualitative results.
\section{More Technical Details}
\subsection{Temporal Interaction (TI): Memory retrieval}

As stated in the paper, we use a memory span $S = 30$ and the total number of cached frame features passed to the TI module is $2S + 1$ (features of the current frame and 30 frames from each side of the current frame). Inspired by \cite{NIPS2015_8fb21ee7} and \cite{tang2020asynchronous}, we keep a cached memory pool $\mathcal{M}$ with temporal person features, and at the end of each iteration, we update the cached person features. This dynamic approach does not increase the computational cost as the number of frame increases. It also allows us to train the model end-to-end since we don't need to pre-train and freeze memory features as is the case with \cite{pan2021actor} and \cite{lin2017feature}. \\

\subsection{How we chose QKV for the interaction modules}
By definition of Cross Attention, the query should stem from the main features and the key and value from the helping features. In our RGB stream, we aim to model interaction between the target actor and the supporting actors (when available) and the context (object and hands). Therefore, the target person is constitutes our main feature. The other features serving as key and and value help us find better representations for the person features (query). Specifically, the hands interaction sub-network outputs better person features after looking at how the target person interacts with the hand features. \textit{Theoretically, the features that will be passed to the final classifier should be the query}. \\

\subsection{Intra-Modality Aggregation} 
\label{subsec:ima}
The intuition for the intra-modality aggregation components $z_r$ and  $z_p$ is to filter and pass only the best of the previously computed features (including the previous block's output and all other interaction blocks) to the next interaction module. We experimented with channel-wise concatenation followed by convolution, feature average, and weighted sum to achieve this. The first is more computationally expensive (Convolution on a very high dimensional vector); the second is straightforward but assumes every feature is equally important. Therefore, we chose to use the weighted sum approach. The only feature fusion mechanism we found in the literature is channel-wise concatenation followed by PCA. However, it suffers from the same shortcomings as the concatenation + convolution approach.
Many initialization schemes came to mind, especially Kaiming and Xavier's initializations. However, they do not work well for this task because they would magnify or reduce the magnitude of our input instead of letting the model learn as it sees fit. We ruled out random initialization since we want every feature to be equally weighted initially. Learning from scratch usually requires small weights, therefore we tried $-k\times ln(10)$ with $k$ taking values $2, 5, 7$, then settled for $k = 5$ (initial weights $\approx{-11.5}$).
\subsection{Implementation details for AVA and MultiSports}
We train AVA with two different activation functions for the classifier. For pose actions, we use a softmax classifier since there can only be one pose for each person. AVA is designed such that one person can have up to three of both person interaction and object interaction labels. Therefore, we use a sigmoid activation for person-interaction-related classes and another sigmoid for object-interaction-related ones. The network is trained for 110k iterations, with the first 2k iterations serving as linear warmup. We use a base learning rate of 0.0004, decreasing by a factor of 10 at iterations 70k and 90k. We use the SGD optimizer and a batch size of 16 to train the model on 8 GPUs. At inference, we predict action labels for human bounding boxes detected by the Faster-RCNN \cite{ren2015faster} instance detector with a confidence threshold of 0.8. We follow the same configurations as AVA to train MultiSports, except for the activation functions. MultiSports does not have overlapping actions; therefore, we only use softmax activation for the classifier.

\section{Extended Qualitative Results}
\begin{figure}[ht]
\centering 
    \includegraphics[width=.9\linewidth]{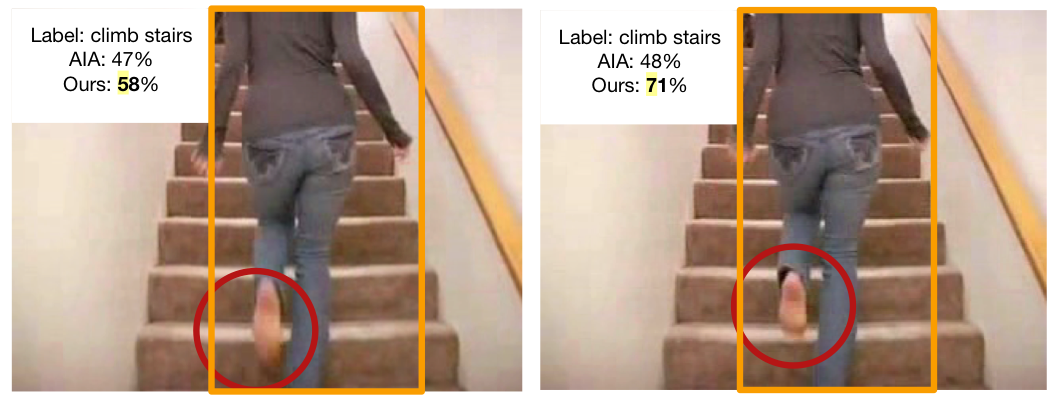}
    \caption{\textbf{Temporal Interaction.} Even though the encircled part is the only noticeable difference between frames $t$(left) and $t+1$(right), our model upgrades the confidence score, as it progresses through the video.}
    \label{fig:progression}
\end{figure}

All of the qualitative results presented here are based on frames extracted from the J-HMDB dataset. \textit{Ours} refers to our proposed \textbf{HIT} network, and AIA is an interaction modeling framework  presented in \cite{tang2020asynchronous}. For a fair comparison, we implement AIA on top of the same backbone we use for our model. 
With Figure \ref{fig:progression}, we want to illustrate how our interaction framework compares to AIA \cite{tang2020asynchronous} in terms of action sequence detection. As the subject is climbing the stairs, her action becomes more an more apparent to our model and the confidence score given to the action sharply increases. However, AIA's confidence score stays stagnant. Such a result highlights the strength of our temporal interaction modeling approach.

\begin{figure*}[t]
\centering
        \begin{subfigure}[t]{.3\textwidth}
            \centering
            \captionsetup{justification=centering}
            \includegraphics[width=.9\linewidth]{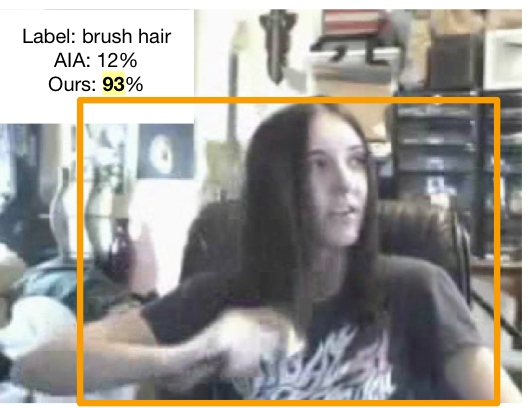}
            \caption{AIA miclassifies the action in this frame as ``wave".}
            \label{fig:sfig1_sup}
        \end{subfigure}%
        \begin{subfigure}[t]{.3\textwidth}
            \centering
            \includegraphics[width=.9\linewidth]{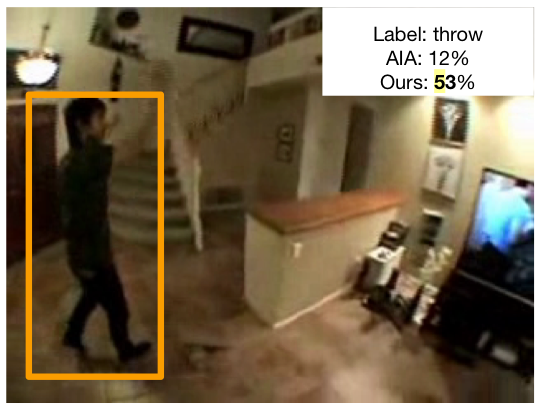}
            \caption{Frame with the action class ``Throw" where the object being thrown is undetected.}
            \label{fig:sfig2_sup}
        \end{subfigure}
        \begin{subfigure}[t]{.3\textwidth}
            \centering
            \includegraphics[width=.9\linewidth]{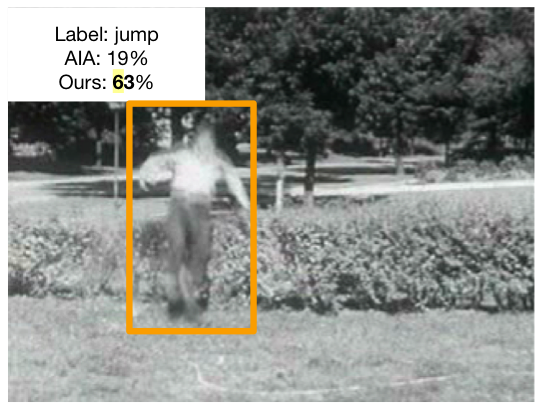}
            \caption{Blurry frame, typical of the action ``jump". The second most challenging class in J-HMDB.}
            \label{fig:sfig3_sup}
        \end{subfigure}
        
        \begin{subfigure}[t]{.3\textwidth}
            \centering
            \includegraphics[width=.9\linewidth]{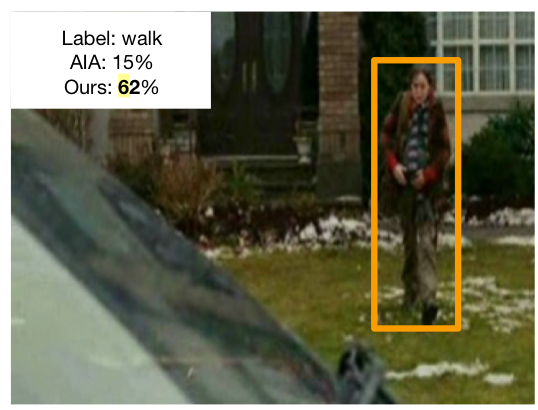}
            \caption{A blurry frame but with clear pose signature.}
            \label{fig:sfig4_sup}
        \end{subfigure}
        \begin{subfigure}[t]{.3\textwidth}
            \centering
            \includegraphics[width=.9\linewidth]{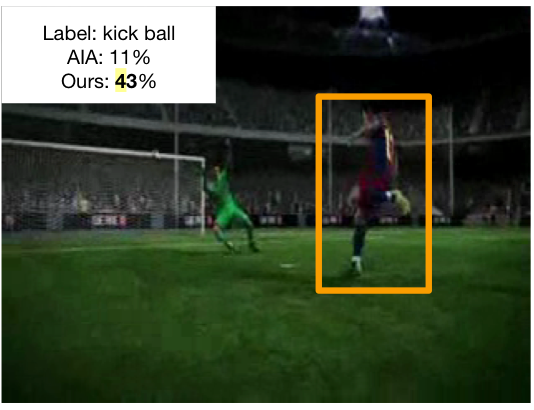}
            \caption{Frame from a video game. AIA evenly distributes the score between many different classes.}
            \label{fig:sfig5_sup}
        \end{subfigure}
        \begin{subfigure}[t]{.3\textwidth}
            \centering
            \includegraphics[width=.9\linewidth]{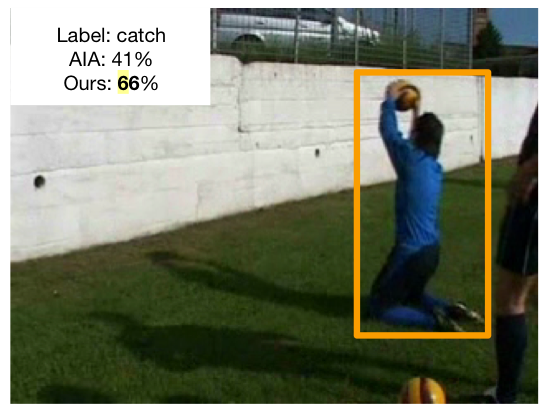}
            \caption{``catch" is strongly hand-related.}
            \label{fig:sfig6_sup}
        \end{subfigure}

        \caption{\textbf{More Qualitative results.} Comparison with AIA}
        \label{fig:qualitative1_sup}
\end{figure*}
\begin{figure*}[t]
\centering 
    \includegraphics[width=\linewidth]{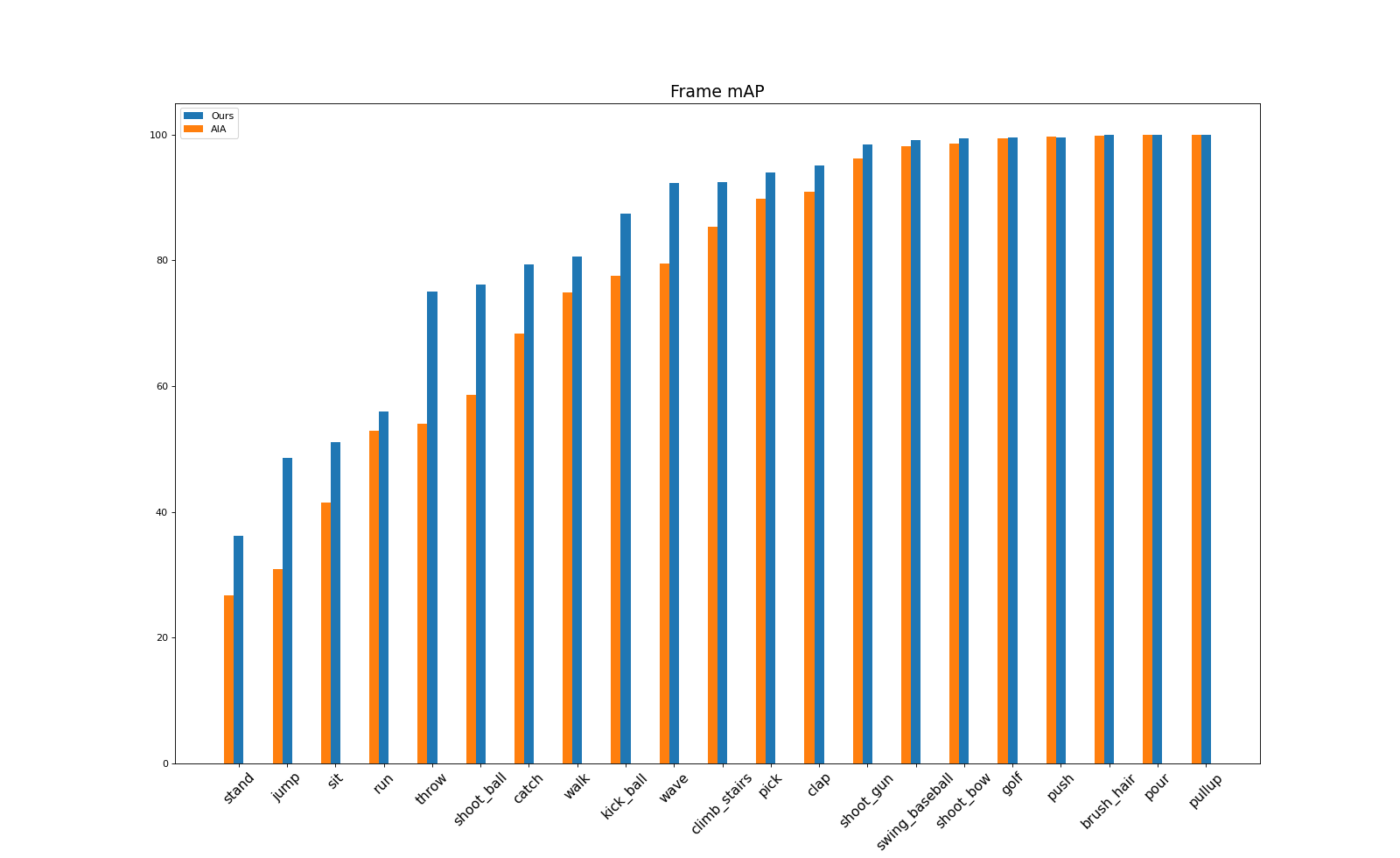}
    \caption{\textbf{Per-class frame mAP comparison with AIA.}}
    \label{fig:comparison_graph}
\end{figure*}

In Figure \ref{fig:sfig1_sup}, AIA misclassifies ``brush hair" as ``wave". Looking at the whole video, any human would detect that the subject is brushing their hair. However, it's not clear for that particular frame. Our model, with strong temporal support and atomic interaction modeling, is able to correctly detect this action. Figure \ref{fig:sfig2_sup} is challenging since the ``throw" action that is happening here is competing with the actions ``stand". Still, our model gives the bulk of the confidence score to the correct action label. Figure \ref{fig:sfig3_sup}  is blurry, just like many of the frames that fall within the ``jump" class. Pose features comes in handy for such actions. We can also see from Figure \ref{fig:comparison_graph} that for that particular action class, our method significantly outperforms AIA. Figures \ref{fig:sfig4_sup} to \ref{fig:sfig6_sup} corroborate the arguments that pose interaction (\ref{fig:sfig4_sup}, \ref{fig:sfig5_sup}) and hand interaction (\ref{fig:sfig6_sup}) are essential for accurate atomic action detection.

Figure \ref{fig:comparison_graph} is a detailed per-class comparison table between our method and AIA. Our method significantly outperforms AIA on challenging hand-related classes such as ``throw", ``catch" and ``wave". It also registers strong performance against classes with fast movements (``jump", ``run"). 
Overall, our HIT network outperforms AIA on every class, providing a significant upgrade on current interaction frameworks used in the literature.

\begin{figure*}[ht]
\centering
        \begin{subfigure}[t]{.45\textwidth}
            \centering
            \captionsetup{justification=justified, singlelinecheck=on}
            \includegraphics[width=.9\linewidth]{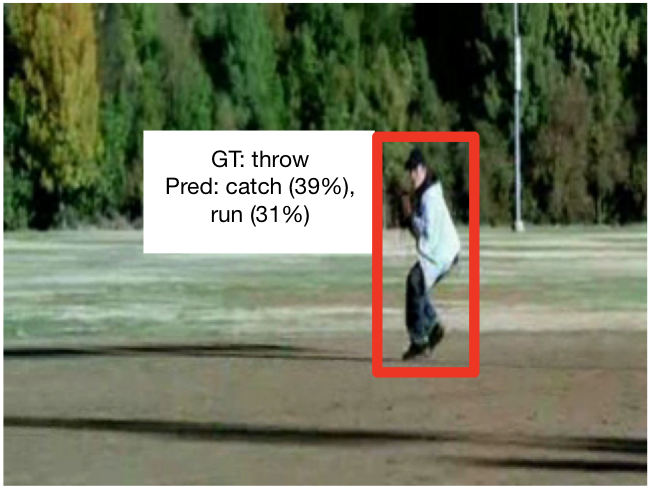}
            \caption{Confusion between the classes ``throw" and ``catch".}
            \label{fig:fail_throw}
        \end{subfigure}\hfill
        \begin{subfigure}[t]{.45\textwidth}
            \centering
            \includegraphics[width=.9\linewidth]{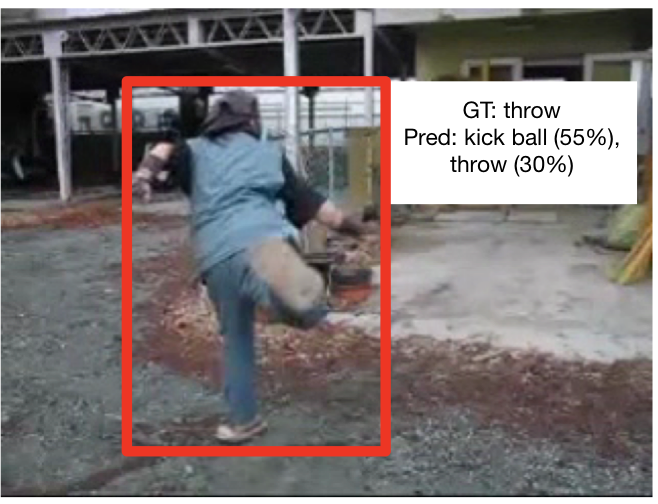}
            \caption{Confusion between the classes ``throw" and ``kick ball".}
            \label{fig:fail_throw1}
        \end{subfigure}\hfill
        
        \begin{subfigure}[t]{.45\textwidth}
            \centering
            \includegraphics[width=.9\linewidth]{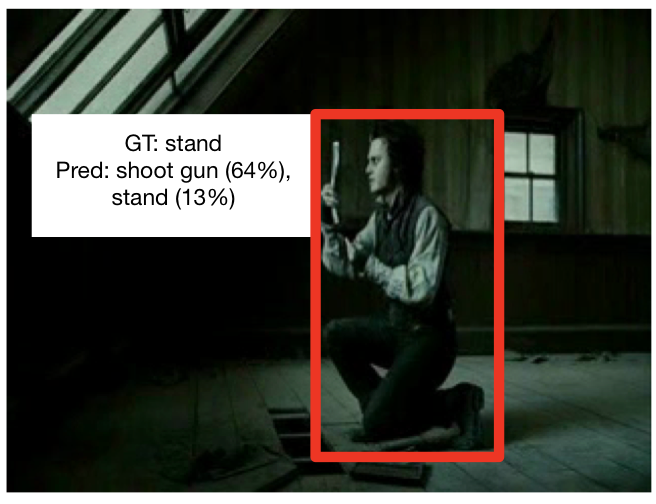}
            \caption{The ground-truth is incorrect in this case.}
            \label{fig:fail_stand}
        \end{subfigure}\hfill
        \begin{subfigure}[t]{.45\textwidth}
            \centering
            \includegraphics[width=.9\linewidth]{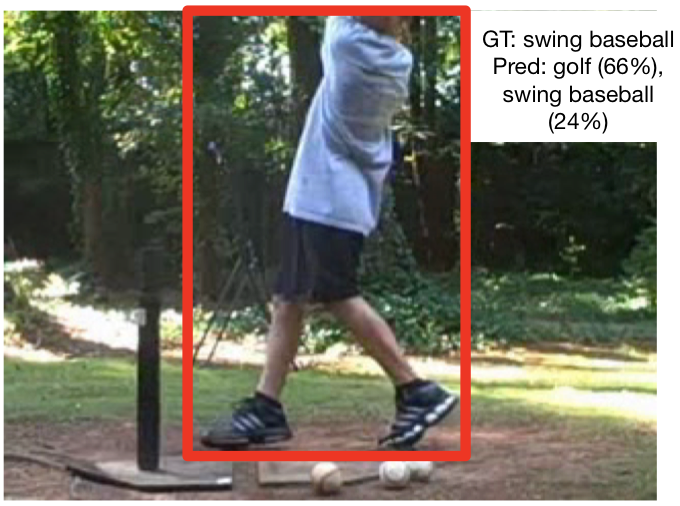}
            \caption{Due to the partial occlusion, ``swing baseball" and ``golf" look almost identical.}
            \label{fig:fail_swing}
        \end{subfigure}\hfill

        \caption{\textbf{Failure cases.} Most failure cases are due to similar-looking classes (a and b), incorrectly labeled classes (c), and partial occlusion (d).}
        \label{fig:failure_cases}
\end{figure*}

\subsection{Failure cases}
Most of the failure cases for our method fall into one of three categories. The first one is similar-looking classes. Classes such as ``throw" and ``catch" usually share the same pose signature and are visually identical. In Figures \ref{fig:fail_throw} and \ref{fig:fail_throw1}, we see how close the class ``throw" can be to the class ``catch" and ``kick ball", respectively. Looking at these frames, any human could misclassify these actions. For these kinds of classes, memory modeling could help. However, it is not guaranteed to work every time. To clear the ``throw"-``catch" confusion, the question the memory features has to answer is the following: is the object coming from or heading to the person's hands? The second common failure has to do with incorrectly labeled classes. For instance, a person standing up is not a person who stands. Someone might kneel while standing up, as shown in Figure \ref{fig:fail_stand}. However, for the J-HMDB dataset, the label is the same throughout the video. Therefore, the act of kneeling while standing up is considered ``stand" and should be classified as such even though it is visually wrong. The third category most methods struggle with is occlusion (see Figure \ref{fig:fail_swing}). Playing golf and swinging a baseball are almost identical, with the key difference being the object the actor is using. Is it a baseball bat or a golf club? In this case, however, the object is occluded. Therefore, the model finds it difficult to differentiate between ``golf" and ``swing baseball."

How should we then go about solving these issues? In our case, we try and aggregate as much information as possible. However, having so much information is costly. The best answer to these problems would be better temporal support, but that would raise another question: how do we define ``better temporal support"?. While some might advocate for more extended temporal support, it would increase the computation overhead while not necessarily translating into higher detection accuracy. Some actions need long temporal support, some need very little, and others need none; therefore, deciding how much memory to keep around is challenging. And if we keep a longer memory span, the need to compress the feature would be more pressing, and most existing compression methods are lossy.

\section{Complexity analysis}
We perform a mini-experiment to determine the FPS of our model with different settings. We could not compare with other frameworks since they do not report these numbers. The backbone has a FPS of around 16, and paired with AIA \cite{tang2020asynchronous}, it is reduced to 13.66. Our model is marginally more complex than AIA due to the added features. However, it can be justified by the considerable improvements in accuracy.
\begin{table}[ht]
 \centering
\begin{tabular}{l|c|c}
\toprule
    Model & frame@0.5 & FPS \\
    \toprule
    Backbone & 58.85 & \textbf{15.97}  \\
    \midrule
    Backbone + AIA\cite{tang2020asynchronous} & 77.25 & 13.66 \\
    Backbone + Ours & \textbf{83.81} & 12.37 \\
    \bottomrule
  \end{tabular}\hfill%
  \caption{\textbf{Performance and FPS comparison between the video backbone, AIA and our framework.}}
    \label{tab:jhmdb_sup}
\end{table}

{\small
\bibliographystyle{ieee_fullname}
\bibliography{egbib}
}

\end{document}